\documentclass[acmsmall]{acmart}

\AtBeginDocument{%
  \providecommand\BibTeX{{%
    \normalfont B\kern-0.5em{\scshape i\kern-0.25em b}\kern-0.8em\TeX}}}

\setcopyright{acmlicensed}
\acmJournal{TALLIP}
\acmYear{2022} \acmVolume{1} \acmNumber{1} \acmArticle{1} \acmMonth{1} \acmPrice{15.00}\acmDOI{10.1145/3546190}

\usepackage{booktabs}
\usepackage{multirow}
\usepackage{algorithm}
\usepackage{amsmath}
\usepackage{algorithmicx}
\usepackage{algpseudocode}
\usepackage{subfigure}

\begin{document}

\title{An Understanding-Oriented  Robust Machine Reading Comprehension Model}


\author{Feiliang Ren}
\email{renfeiliang@cse.neu.edu.cn}
\orcid{0000-0001-6824-1191}
\affiliation{%
  \institution{Northeastern University}
  \streetaddress{11 Wenhua Rd}
  \city{Heping Qu}
  \state{Shenyang Shi}
  \country{China}}

\author{Yongkang Liu}
\orcid{0000-0003-3098-0225}
\affiliation{%
	\institution{Northeastern University}
	\country{China}
}
\author{Bochao Li}
\orcid{0000-0003-2897-3886}
\affiliation{%
	\institution{Northeastern University}
	\country{China}}
\author{Shilei Liu}
\orcid{0000-0003-2976-6256}
\affiliation{%
	\institution{Northeastern University}\country{China}}
\authornote{These authors contribute equally to this research and are listed randomly.}
\author{Bingchao Wang}
\orcid{0000-0002-3528-773X}
\affiliation{%
	\institution{Northeastern University}
	\country{China}}
\authornotemark[1]
\author{Jiaqi Wang}
\orcid{0000-0001-9306-8757}
\affiliation{%
	\institution{Northeastern University}
	\country{China}}
\authornotemark[1]
\author{Chunchao Liu}
\orcid{0000-0002-0028-8425}
\affiliation{%
	\institution{Northeastern University}
	\country{China}}
\authornotemark[1]
\author{Qi Ma}
\orcid{0000-0001-9548-1350}
\affiliation{%
	\institution{Northeastern University}
	\country{China}}
\authornotemark[1]

\renewcommand{\shortauthors}{Ren, et al.}

\begin{abstract}
Although  existing machine reading comprehension models are making rapid progress on many datasets, they are far from robust. In this paper, we propose an understanding-oriented  machine reading comprehension model to address three kinds of robustness issues, which are over sensitivity, over stability and generalization. Specifically, we first use a natural language inference module to help the model understand the accurate semantic meanings of input questions so as to  address the issues of  over sensitivity and over stability. Then  in the machine reading comprehension module, we propose a memory-guided multi-head attention method  that can further well understand the semantic meanings of input questions and passages.  Third, we propose a multi-language learning mechanism to address the issue of generalization.  Finally, these modules are  integrated with a multi-task learning based method.  We evaluate our model on  three benchmark datasets that are designed to measure models' robustness, including DuReader (robust) and two SQuAD-related datasets.  Extensive experiments show that our model can well address the mentioned three kinds of robustness issues. And it achieves much better results than the compared state-of-the-art models on all these datasets under different evaluation metrics, even under some extreme and unfair evaluations. The source code of our work is available at: {https://github.com/neukg/RobustMRC}.

%To address these two issues, we propose a deep understanding based MRC model that is suitable to the scenarios like the multi-document MRC task and the low-cost hardware environments. Specifically, it has three cascaded deep understanding modules which are designed to understand the accurate semantics of words, the interactions between the input  question and documents, and the supporting cues for  the correct answer. We evaluate our model on  two large scale benchmark datasets, TriviaQA Web and  DuReader. Extensive experiments show that our model  achieves state-of-the-art results on both  datasets.

%On the other hand we notice that most existing multi-document machine reading comprehension models mainly focus on understanding the interactions between the input question and  documents, but ignore following two kinds of understandings. First, to  understand the semantics of words in the input question and documents  from the perspective of each other. Second, to understand the supporting cues for a correct answer from the perspective of intra-document and inter-documents. These two kinds of ignoring would make the models oversee some important information that might be helpful for finding correct answers. To address these issues, we propose a deep understanding based model for multi-document machine reading comprehension. 

\end{abstract}





\maketitle

\section{Introduction}

Machine reading comprehension (MRC) aims to answer questions by reading given passages (or documents). 
It is considered one of the core abilities of artificial intelligence (AI) and  the foundation of many AI-related applications like next-generation search engines and conversational agents.

%~\cite{joshi2017triviaqa,clark2018simple,hu2019retrieve,hu2019read,gan2019improving}
At present, MRC is achieving more and more research attention and lots of novel models have been proposed. On some benchmark datasets like SQuAD~\cite{rajpurkar2018know} and MS MARCO~\cite{nguyen2017ms}, some models like {BERT}~\cite{devlin2018bert} have achieved higher performance than human. However, recent work \cite{jia2017adversarial,gan2019improving,tang-etal-2021-dureader,zhou2020robust,liu2020a,wu2020improving,si2020benchmarking} shows  that current MRC test sets tend to overestimate an MRC model's true ability to unseen data due to the following reason:   the test set on which an MRC model  evaluated  is typically randomly selected from the whole set of data collected and thus follows the same distributions as the training and development sets, while in real world, it is impossible to ask the unseen data follow such known distributions. Thus it is very necessary to evaluate MRC models on some unseen test data to reveal their robustness.  

\begin{figure*}[t]%%图
	\centering  
	\includegraphics[width=0.9\linewidth]{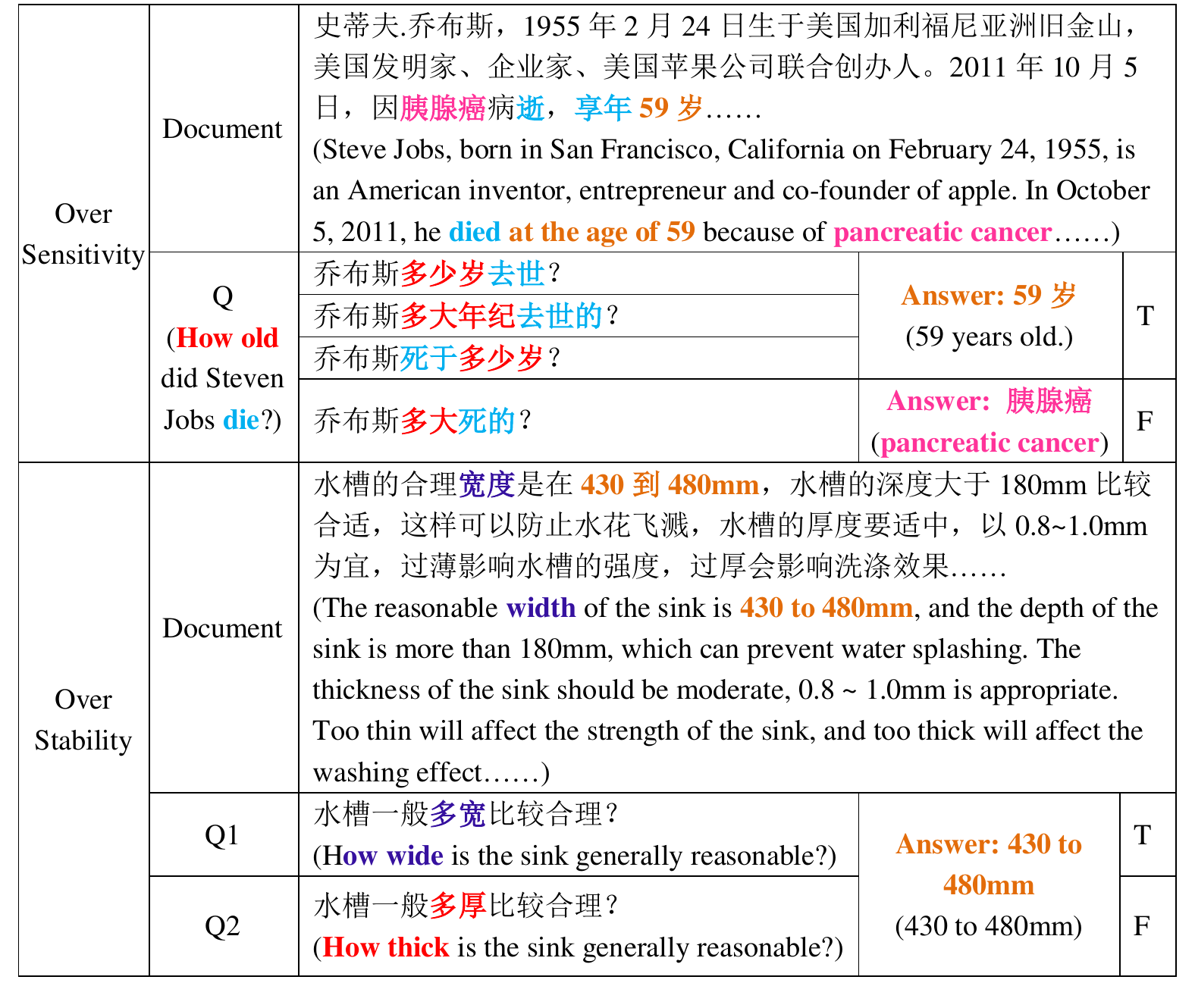} 
	\caption{Examples of over sensitivity and over stability (extracted from DuReader (robust)). The words with the same color have the same meaning. All the answers are generated by BERT (large). }
	\label{fig:example}
\end{figure*}

In this study, we focus on following three kinds of robustness issues that are defined by \cite{tang-etal-2021-dureader}. (i) Over sensitivity issue which refers to semantically invariant text perturbations  cause a models' prediction to change when it should not; (ii) Over stability issue which refers to input text is meaningfully changed but the model’s prediction does not, even though it should; (iii) Generalization issue which refers to models usually perform well on in-domain test sets yet perform poorly on out-of-domain test sets.   All these   issues are widely existed in real world, and they will lead to a significant decrease in performance for most of existing state-of-the-art MRC models~\cite{jia2017adversarial,gan2019improving,welbl2020undersensitivity,tang-etal-2021-dureader,liu2020a}. Fig. \ref{fig:example} shows two examples about  the issues of  over sensitivity and over stability respectively, both of which  are  extracted from the DuReader (robust) dataset \cite{tang-etal-2021-dureader} and   all the questions in them  are really asked by users in the \emph{Baidu} search engine. 

Existing methods  address above issues mainly with  a kind of data augmentation based methods  or a kind of adversarial training based methods.  For example, Gan and Ng\shortcite{gan2019improving}  use a neural paraphrasing model to generate multiple paraphrased questions for a given source question that is paired with a set of paraphrase suggestions. Then an MRC model is retrained on the training set where the paraphrased samples are integrated. 
However, neither of these two kinds of  methods can truly solve the mentioned issues. Essentially, they still focus on the effort of making a model ``\emph{see}" as many samples as possible so that the model can make decisions based on the ``\emph{saw}" knowledge. But both the paraphrased and adversarial questions are ``\emph{generated}", so it is very possible that they may not be present in real world. Besides, many real questions could not be fully ``\emph{generated}". % Accordingly, either the risk of overestimating an MRC model or the two issues are  not eliminated fundamentally.  
Taking the questions in  Fig. \ref{fig:example} as examples, for the first one, there are many different ways to express “\emph{how old}" or “\emph{die}” in Chinese. For the second one, there is  only ONE different Chinese CHARACTER in the two questions, but they have completely different semantic meanings. Both  examples are difficult to be addressed by existing data augmentation or adversarial training based methods because of  the following reasons. For the over sensitivity example, there are too many diverse and flexible expression manners for a same meaning question to be enumerated or paraphrased. As for the over stability example, it is impossible to paraphrase it because the aim of data augmentation methods is to generate some   paraphrased questions that have  the same semantic meaning with the source question. And  it is also impossible to handle this over stability example by the adversarial training based methods because these methods  usually use some context words near a wrong answer candidate to \emph{generate} some adversarial examples  based on which the models are trained \cite{gan2019improving}. Thus these methods do not have capabilities to distinguish the slight perturbations between two questions that have a significant high string match similarity. %minor difference between two but here the difference between two questions of this examples would be too minor to be distinguished by the models. %caught. % by an  adversarial training based model. % to imitate the {over stability} questions because .   

In contrast,  human can handle 	all above  three kinds of robustness issues effectively,  the main reason of which is that human can \emph{understand} the semantic meanings of the given text precisely. In fact, the \emph{understanding} capability  is also the key of  solving these diverse kinds of robustness issues~\cite{jia2017adversarial}. Inspired by this, we propose an  \emph{understanding-oriented} MRC model that can address the mentioned three kinds of robustness issues well. First, we view both  the issues of over sensitivity and over stability  as a \emph{semantic meaning understanding} problem, which requires the MRC model have the ability of  distinguishing the semantic meanings of a question and its paraphrased expressions that may have similar or dissimilar semantic meanings with the source question.  To this end,  we introduce a \emph{nature language inference} ({NLI})~\cite{2006The}  module to judge whether two input sentences have the same semantic meaning.  Second, in the MRC module, we propose a  \emph{memory-guided multi-head attention}  method that can  better \emph{understand} the interactions between   questions and passages. Third, we propose a \emph{multi-language  learning} mechanism to prevent the model from over-fitting in-domain data and enhance the generalization ability of the model. We introduce several language-specific MRC datasets to train the model together, which makes the distributions of training set and test set be completely different. During training,   each dataset can be viewed as  an adversarial dataset of others. Accordingly, none of a dataset can dominant the training process and the model will be pushed to learn more generalized knowledge for predictions.  And these modules are   jointly trained with a multi-task learning  manner. 

We evaluate the proposed method on three benchmark MRC datasets, including DuReader (robust) \cite{tang-etal-2021-dureader} and two SQuAD-related datasets \cite{gan2019improving}. All of these datasets are designed to  measure the capabilities of an MRC model for addressing the issues of over sensitivity, over stability and generalization. Extensive experiments  show that our model achieves  very competitive results on all of these datasets.
On DuReader (robust), it  outperforms the compared strong baselines by a large margin and ranks No.3 on the final test set leader-board. On the other two SQuAD-related datasets, it achieves much competitive results even under two kinds of  extreme and unfair evaluations. %conditions. %, which demonstrates its strong   generalization ability. %   than the state-of-the-art baselines.  

%Most existing MRC models pay less attention to the following three issuses. 
%First, the over sensitivity issue. A model is considered to be over-sensitive if it fails to give the same and correct answer to the paraphrased questions.
%Second, the over stability issue. A model is considered to be over-stable if it strongly depends on lexical matching and is not able to distinguish the missing leading sentences and the ones containing the reference answers.
%Third, the generalization issue.  A model is considered bo be with weak generalization capability if it suffers significant performance drop in an out-of-domain test set. 

%These instructions are for authors submitting papers to the NAACL-HLT 2021 conference using \LaTeX. They are not self-contained. All authors must follow the general instructions for *ACL proceedings,\footnote{\url{http://acl-org.github.io/ACLPUB/formatting.html}} as well as the newly introduced formatting guidelines for an optional ethics/broader impact section (see the conference website at \url{https://2021.naacl.org/}).  This document contains additional instructions for the \LaTeX{} style files.

%The templates include the \LaTeX{} source of this document (\texttt{naacl2021.tex}),the \LaTeX{} style file used to format it (\texttt{naacl2021.sty}), an ACL bibliography style (\texttt{acl\_natbib.bst}), an example bibliography (\texttt{custom.bib}), and the bibliography for the ACL Anthology (\texttt{anthology.bib}).

\section{Related Work}
\textbf{Common MRC Research} In the early study of MRC,  researchers pay much attention to design diverse attention  methods to  mine the interactions between  questions and passages. These interactions have been proven to be much helpful for improving the performance of an MRC model.  BiDAF \cite{seo2016bidirectional} is one of the most representative work, where the authors design  a Context-to-query and Query-to-context bi-directional attention method.  \cite{yu2018qanet} and \cite{clark2018simple} also use a BiDAF-style attention method. Besides, researchers also propose  many other kinds of attention methods. For example, \cite{cui2017attention} designs an attention-over-attention model that uses a 2-dimension similarity matrix between the question and the context words to compute the weighted   query-to-context attention. \cite{wang2018multi} propose a multi-granularity hierarchical attention method.  \cite{hu2019retrieve}  use the self-attention based method. 

Recently, there are two kinds of research lines that are dominant in the  MRC task, both of which achieve competitive results  on many benchmark MRC datasets.  

The first one is to imitate some reading patterns used by human when designing an MRC model.  For example, \cite{sun2019improving} explicitly use three human’s reading strategies in their MRC model, including: (1) back and forth reading, (2) highlighting, and (3) self-assessment.
\cite{wang2018multipassage} imitate human's following reading pattern: first scans through the whole passage; then with the question in mind, detects a rough answer span; finally, come back to the question and select the best answer. \cite{liu2018stochastic} design their MRC model by simulating human's multi-step reasoning pattern: human  often re-read and re-digest given passages many times before a final answer is found. \cite{wang2018joint} use an extract-then-select reading strategy.  They further regard the candidate extraction as a latent variable and train the two-stage process jointly with reinforcement learning.  \cite{peng2020bi} design their MRC model by  simulating two ways of human thinking when answering questions, including reverse thinking and inertial thinking.    \cite{zhang2021retrospective} imitate human's \emph{``read + verify"}   reading pattern: first to read through the full passage along with the question and grasp the general idea, then re-read the full text and verify the answer. Some other researchers~\cite{hu2019read,clark2018simple,yan2019a,wang2018multi} also imitate human's this \emph{``read + verify"}   reading pattern. There are other kinds of human reading patterns imitated. For example, ~\cite{tian2020scene} imitate the pattern of restoring a scene according to the text, ~\cite{malmaud2020bridging} imitate the pattern of human gaze during reading comprehension, and \cite{chen2020multi} imitate the pattern of  tactical comparing and reasoning over candidates while choosing the best answer, etc.

The second one is to use diverse large-scale pretrained language models like {BERT}~\cite{devlin2018bert} and lots of its variants including  {XLNet}~\cite{yang2019xlnet}, RoBERTa~\cite{liu2019roberta}, and ALBERT~\cite{lan2020albert}, etc. These language models  have a strong capacity for capturing the contextualized sentence-level language representations \cite{zhang2021retrospective}. With these language models, researchers can design an MRC model very easily because the language models can be used either as MRC models themselves, or as the encoder part of an MRC model and  researchers only need to focus on  designing  the decoder part.    For example, lots of recent MRC models \cite{gong2020recurrent,zheng2020document,luo2020map,long2020synonym,banerjee-etal-2021-self,li2020towards,zhang2020learn,guo2020incorporating,guo2020a,li2020mrc,huang2020nut,chen2020forcereader} only consist of   a language model based encoder module and a careful designed decoder module. %These language model based MRC model

%However, there is a fatal deficiency for these language models. First, except XLNet,  BERT and its other variants (ALBERT, RoBERTa, etc) are autoencoding based models, which  limits input size of  512 TOKENS~\cite{zemlyanskiy-etal-2021-readtwice,gong2020recurrent}. This restriction has no effect on most \emph{AI}-related   applications and most of single-document MRC tasks, but for an multi-document MRC dataset like DuReader or TriviaQA Web, this restriction will make most correct answers be excluded from the input documents even after a carefully designed data selection module. As for XLNet, it is an autoregressive based model, and can handle long text theoretically. However, it is an uni-directional model which can make predictions based on  forward information only, and can not use the backward information. 

\noindent\textbf{Robust MRC Research}  \cite{jia2017adversarial} explore the over sensitivity and over stability issues by testing whether an MRC model can answer paraphrased questions  that contain adversarial sentences. {Recently, \cite{naplava-etal-2021-understanding} extensively evaluated several state-of-the-art AI-related downstream systems including some MRC models with their robustness to input noise.} Their experiments show that there is no published open-source models which are robust to the addition of adversarial sentences. Thus, the robustness issues in MRC are attracting more and more research attentions, and lots of novel methods have been proposed. Generally, these existing methods can be  classified into following two kinds.

The first kind is the data augmentation based methods that address the robustness issues by training a model with additional careful generated training data.  For example, \cite{wang2018robust} augment the training datasets by incorporating
some adversarial examples, and then the MRC model is trained  on the augmented dataset.  \cite{welbl2020undersensitivity} investigate data augmentation and adversarial training as defenses. \cite{liu2020a}  propose a model-driven approach to generate adversarial examples that can attack given MRC models. Then they retrain and strengthen the MRC model by using the generated adversarial examples. {\cite{li-etal-2021-select-one}  introduce a new knowledge distillation method by taking advantage of data augmentation and progressive training on a wide range of AI-related applications including the MRC task.   \cite{shinoda-etal-2021-improving} focus on question-answer pair generation to mitigate the robustness issue. But different from most existing methods that aim to improve the quality of synthetic examples, they try to generate multiple diverse question-answer pairs to mitigate the sparsity of training sets so as to improve the robustness of a model.}  Usually, this kind of methods are simple and effective. However,  \cite{liu2020a} point out that the   augmented datasets are often capable of simulating the known types of adversarial examples, while ignoring other unobserved types. \cite{gan2019improving} further point out that the main deficiency in the data augmentation based methods is that the adversarial examples created are unnatural and not expected to be present in real world.  {\cite{rosenberg-etal-2021-vqa} draw similar conclusion that data augmentation based methods cannot address the  	robustness issues  effectively.}
%They  explore the robustness issue by creating two test sets consisting of paraphrased SQuAD questions. These two test sets are designed to test QA models' over sensitivity and over stability respectively.  

The second kind is the adversarial training based methods that  explore to design better MRC models to improve models' robustness. For example,  \cite{liu2018stochastic} average multi-predictions to improve the model's robustness. \cite{min2018efficient} notice that most questions can be answered by using only a few sentences and without the consideration of context over entire passage, then they design a sentence selector to select the minimal set of sentences  to the MRC model  to answer a question. Their method  reduces the risk of  adversarial attacks by reducing passage length, which is proven to be robust to adversarial inputs.
\cite{2021Ensemble} investigate the effect of ensemble learning approach to improve the generalization of MRC models. After separately training several base models with different structures on different datasets, these base models are ensembled by using weighting and stacking approaches in probabilistic and non-probabilistic settings.  Conversely,  \cite{hu2018attention} train a robust single model based on ensemble ones through distillation training approach. They first apply the standard knowledge distillation to mimick output distributions of answer boundaries from an ensemble model, then  propose two distillation approaches to further transfer knowledge between the teacher model and the student model.  {\cite{bartolo-etal-2021-improving} use synthetic adversarial data 	generation to make MRC models 	more robust.}

Besides, some researchers also explore the methods of introducing extra knowledge to address the robustness issues.  For example, \cite{wang2019explicit} address the robustness issues by proposing a data enrichment method that uses WordNet to extract inter-word semantic connections as general knowledge from each given passage-question pair, then they  uses the  extracted  knowledge to assist the attention mechanism in their MRC model. \cite{zhou2020robust}  address the over confidence issue and the over sensitivity issue   simultaneously with the help of external linguistic knowledge. Specifically, they first incorporate external knowledge to impose different linguistic constraints (entity constraint, lexical constraint, and predicate constraint), and then regularize MRC models through posterior regularization. Linguistic constraints induce more reasonable predictions for both semantic different and semantic equivalent adversarial examples, and posterior regularization provides an effective mechanism to incorporate these constraints. \cite{wu2020improving} address the robust issues from two aspects. First, they enhance the representation of the model by leveraging hierarchical knowledge from external knowledge bases. Second, they introduce an auxiliary unanswerability prediction module  and perform  multi-task learning with a span prediction task. % for which a model erroneously predicts the same answer, and with even higher probability.% They experiment with data augmentation and adversarial training as defences.
{Some researchers explore to use auxiliary tasks to address the robustness issues. For example, \cite{chen-durrett-2021-robust} propose a MRC model that  through sub-part alignment, their basic idea is that if every aspect of the question is well supported by the answer context, then the answer produced should be
trustable; if not, they suspect that the model is making an incorrect prediction. And the sub-parts used are predicates and arguments from the results of a \emph{Semantic Role Labeling} task.}

% passages. but  pays more attention to  precisely \emph{understanding} the semantic meanings of both questions and passages passages %just as analysed above, these existing models still focus on the effort of making a model ``\emph{see}" more samples and then make decisions based on the ``\emph{saw}" knowledge. %Thus they .
%From these work we can conclude that prior work highlighted the severity of the robust  issue but this issue is still  not  achieved enough attention.

{\textbf{Natural Language Inference} Given two sentences, often called as a premise and a hypothesis respectively, 	Natural Language Inference (NLI),  also known as \emph{Recognizing Textual Entailment}~\cite{2006The}, is usually defined as  the task of 	determining whether the premise has a relation of \emph{entailment}, \emph{neutral}, or \emph{contradiction} with the hypothesis~\cite{2006The,zhou-bansal-2020-towards,zylberajch-etal-2021-hildif}. According to ~\cite{2006The}, a premise entails a hypothesis if a human reading the premise would infer that the hypothesis is most likely true. For example, given a premise ``\emph{iPhone13 has seen strong sales in China.}'' and a hypothesis ``\emph{Strong sales for iPhone13 in China.}'', their relation should be \emph{entailment}. That is, the given premise entails the hypothesis. In contrast, the relation would be \emph{contradiction} if the hypothesis is ``\emph{iPhone13 is not popular in China}'', and the relation would be \emph{neutral} if the hypothesis is  ``\emph{Strong sales for iPhone13 in Japanese.}''. } 

{Recently, there are many  large-scale	standard datasets  released, like SciTail~\cite{2018SciTaiL}, SNLI~\cite{bowman-etal-2015-large},  Multi-NLI~\cite{williams-etal-2018-broad}, etc. These datasets facilitate the study of NLI greatly, and some  state-of-the-art neural models have achieved very competitive performance  on these datasets~\cite{zhou-bansal-2020-towards,chen-etal-2021-neurallog,belinkov-etal-2019-dont,jiang-etal-2021-alignment,meissner-etal-2021-embracing}. From the definition of NLI  we can see that it is based on (and assumes) common human understanding of language as well as common background knowledge, thus it has been considered by many as an important	evaluation measure for language  understanding \cite{2006The,williams-etal-2018-broad,bowman-etal-2015-large,zylberajch-etal-2021-hildif}. Accordingly, more and more researchers use the NLI task into diverse downstream applications with the expectation that NLI would be useful for these downstream applications. For example, \cite{welleck-etal-2019-dialogue} use the NLI models to improve the consistency of a dialogue model where utterances are re-ranked using a NLI model. \cite{falke-etal-2019-ranking} use the entailment predictions of NLI models to re-rank the generated summaries of some state-of-the-art models. \cite{huang-etal-2021-improving} use the NLI models to improve unsupervised commonsense reasoning. \cite{koreeda-manning-2021-contractnli-dataset} use the NLI models to assist contract review.}

{Like our method, \cite{chen-etal-2021-nli-models} also use an NLI model in their MRC model. But   they aim to train NLI models to evaluate the predicted answers by an MRC model. Specifically, they leverage large pretrained models and recent prior datasets to construct 	powerful question conversion and decontextualization 	modules, which can reformulate question-answer instances as premise-hypothesis pairs with very high reliability. Then, they  combine standard NLI datasets with the NLI examples automatically derived from MRC training data to train the NLI model. }

Essentially, our model makes full use of the advantages in   diverse existing research lines, but  pays more attention to  precisely \emph{understanding} the semantic meanings of  questions and passages. 

\begin{figure*}[t]
	\centering  
	\includegraphics[width=0.99\linewidth]{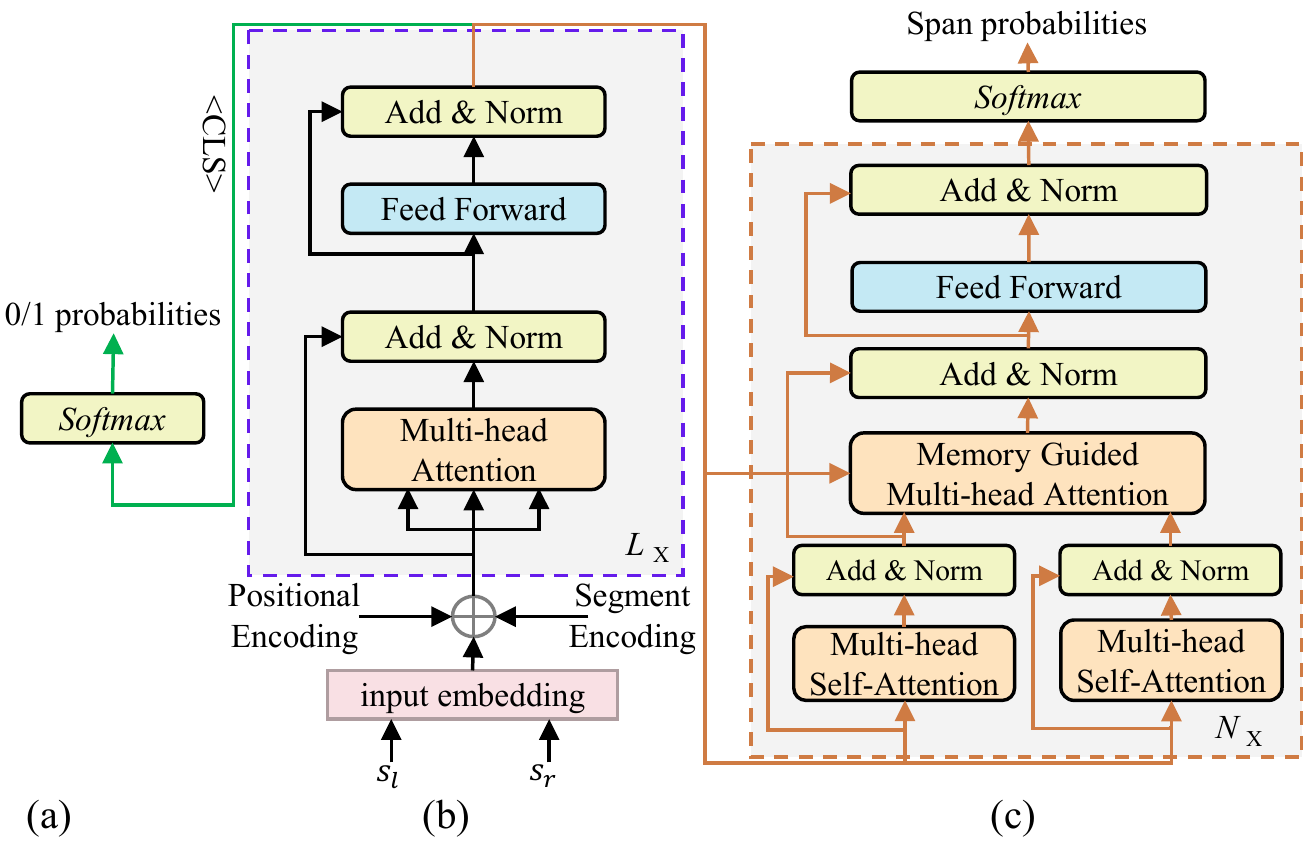}
	\caption{The Architecture of Our Model.} 
	\label{fig:frame}
\end{figure*}

\section{Methodology}

{In this study, both the issues of  {over-sensitivity} and {over-stability} are viewed as a \emph{semantic meaning understanding} problem.  On the other hand, according to the task definition of NLI we can  draw the conclusion  that  if two sentences have the same semantic meaning, they would always be assigned an \emph{entailment} relation. 	Inspired by this,    we convert the traditional NLI task into the task of judging whether two questions have the same semantic meaning.  Concretely, we roughly regard the \emph{entailment} relation between two sentences  as an equivalent alternative for the relation of \emph{having the same semantic meaning}. That is,   we think two sentences  would have an  \emph{entailment} relation if they  have the same semantic meaning, and  they would have a \emph{contradiction} relation if they do not have the same semantic meaning. Furthermore, to make our MRC model be sensitive to subtle semantic changes, the converted {NLI} task is combined with the {MRC} task with a \emph{multi-task learning} framework.} Finally, the architecture of our model is shown in Fig. \ref{fig:frame}. It has three main modules: (i) a   encoder module  (Fig. \ref{fig:frame}(b)); (ii) an MRC module  (Fig. \ref{fig:frame}(c)); (iii) an  {NLI} module (Fig. \ref{fig:frame}(a)). %The NLI module and the MRC module are integrated with a \emph{multi-task and multi-language learning} based method. %Before introducing these modules, we first give some basic notations for   description convenience in subsequent sections. %Concretely, In fact, the rationality of this task conversation lies in the fact that according to the task definition of NLI, a paraphrased sentence of the \emph{premise} would always have the \emph{entailment} relation with the original \emph{premise}. In other words, if two sentences have the same semantic meaning, they would always be assigned an \emph{entailment} relation.}

{For description convenience, here we first give some basic notations about the MRC module. Given a passage with $n$ tokens (denoted as $ p= \{p_i \}_{i=1}^n $) and a question with $m$ tokens (denoted as $ q=\{ q_j \}_{j=1}^m $), an MRC model is to predict an answer \emph{a} that is constrained as a contiguous span in $p$, i.e., $a= \{p_i \}_s^e$, where $s$ and $e$ indicate the beginning and ending positions of the  answer. The training set of an MRC task can be denoted  as $\mathcal{D}^{M}=\{(p^i, q^i, a^i)\}_{i=1}^M$ where $M$ is the number of samples.  In this set, each sample  consists of a passage $p$, a question $q$ and an answer $a$.} %Our MRC model can be formally defined with Equation \ref{eq:mrc}.  
%\begin{equation}
%f_{MRC}(q,p;\theta_S,\theta_M) = \mathop{argmax} \limits_{a^{begin},a^{end}} P\left(a^{begin}|p,q;\theta_S,\theta_M\right) P\left(a^{end}|p,q;\theta_S,\theta_M\right)
%\label{eq:mrc}
%\end{equation}

%In all the subsequent parts of this paper, $\theta_S$ represents the task-independent parameters shared by both MRC and {NLI} tasks. $\theta_M$ and $\theta_N$ are task-dependent parameters for MRC and {NLI} respectively.

\subsection{Shared Encoder Module}
\label{sec:endoder}
In our model, we  use {RoBERTa} \cite{liu2019roberta}  as the shared encoder module  to generate representations for the inputs of both MRC and NLI. This encoder module will output  a   context-aware representation for each token of the input text. Both NLI and MRC take text pairs as input (NLI takes  question pairs as inputs and MRC takes \emph{passage-question} pairs as inputs), so for simplicity, here we use $ S_l$ and $S_r$ to denote the two text  parts of an input text pair and denote the output of this encoder module as $\mathbf{H}$ which  is computed by Equation \ref{eq:h}.
\begin{equation}
\begin{aligned}
\mathbf{H}={\rm RB}([{<}{CLS}{>} \oplus S_l \oplus {<}Sp{>} \oplus S_r \oplus {<}Sp{>}])
\label{eq:h}
\end{aligned}
\end{equation}

where {\rm RB} denotes the RoBERTa model,  {\emph{ <CLS>}} is a padding token,  {\emph{ <Sp>}} is a separator token defined to separate the token sequences of $ S_l$ and $S_r$, and $\oplus$ denotes the concatenation operation.% is a label token defined in almost all language model based applications. 

%~\cite{zhang2021retrospective}. strong capacity for capturing the contextualized sentence-level language representations.
The padded question representation sequence
$\mathbf{H}_Q=\{\boldsymbol{h}^q_1 , ... , \boldsymbol{h}^q_{m}, \boldsymbol{h}_{<Sp>}, $ $ \boldsymbol{0}_1, ... \boldsymbol{0}_{n+1}\}$ and 
the padded passage representation sequence $\mathbf{H}_P=\left\{\boldsymbol{0}_1, ... \boldsymbol{0}_{m+1}, \boldsymbol{h}^p_1 , ... ,\boldsymbol{h}^p_{n}, \boldsymbol{h}_{<Sp>}\right\}$ are used for the subsequent operations in  the MRC module. Here $\boldsymbol{0}_i$ is a padded zero vector whose items are all zeros. Obviously,  after the padding operations, $\mathbf{H}_P$ and $\mathbf{H}_Q$  have the same vector dimension. % after the padding operations. 

Note  the notations in  Fig. \ref{fig:frame}(b) (like \emph{segment encoding}, \emph{positional encoding}, etc)  have completely the same meanings as those defined in the original paper of BERT~\cite{devlin2018bert} or RoBERTa~\cite{liu2019roberta}, so one can read these original papers for detailed information. %about it)

%where {\rm RB} denotes the RoBERTa model, {\emph{ <Sp>}} and {\emph{ <Cs>}} are two placeholders defined to separate the token sequences of the two text segments in an input text pair.  We take $\boldsymbol{h}_{<Cs>}$ (the context-aware representation of  {\emph{ <Cs>}}) as the input for the NLI module, and take $\mathbf{H}_Q=\left\{\boldsymbol{h}^q_1 , ... , \boldsymbol{h}^q_{m}, \boldsymbol{h}_{<Sp>}, \boldsymbol{0}_1, ... \boldsymbol{0}_{n+1}\right\}$ and  $\mathbf{H}_P=\left\{\boldsymbol{0}_1, ... \boldsymbol{0}_{m+1}, \boldsymbol{h}^p_1 , ... ,\boldsymbol{h}^p_{n}, \boldsymbol{h}_{<Sp>}\right\}$ as the input for the MRC module.

\subsection{MRC Module}
%Our MRC module is also based on \emph{RoBERTa} which  has a \emph{Transformer} \cite{vaswani2017attention} based structure. However, MRC is an \emph{understanding} task while the \emph{Transformer} based deocder is more suitable for a \emph{generation} task. A \emph{generation} task only needs to ``\emph{see}" previous text information, while an \emph{understanding} task need to ``\emph{see}" the complete context information.

The structure of our MRC module is shown in  Fig. \ref{fig:frame}(c). It consists of  $N$ identical computation blocks followed by a multi layer perception (MLP) based output layer. Each computation block  has three components: (i) a multi-head self-attention component that is used to compute a kind of self-aware representations  for the input question and passage, with which the model could ``\emph{see}" the entire context information; (ii) a  ``\emph{memory-guided interaction mining}"  component that can mine richful interactions between question and passage; and (iii) a \emph{softmax} based output component. 

Our MRC module is  based on the decoder component in {RoBERTa}, thus as shown in  Fig. \ref{fig:frame}(c), it has a similar structure as that in \emph{Transformer}'s decoder \cite{vaswani2017attention}. The main difference between our model and the original decoder in  \emph{Transformer} lies in the second component.%, which will be described in detail in the subsequent section. Here we omit the descriptions for the first and third components in the computation block  because both of them have the same operations as those defined in  \emph{Transformer}'s decoder \cite{vaswani2017attention}, and readers can find the detailed descriptions of them easily from the original paper of \emph{Transformer}. % for detailed information. 

%It has three main components. First,  we add an \emph{Multi Head Self-Attention} layer to compute a kind of self-aware representations  for the input question and passage, with which the model could ``\emph{see}" the whole context information. Second, we change the original \emph{Multi-Head Attention} component with  an \emph{Mixed Similarity Matrix based Multi Head Attention} component, in which we compute the question-aware passage's representation with a mixed similarity matrix based   method that can mine more interactions between question and passage.  We employ a resudual connection followed by the layer normalization in each of these two component. Finally, there is a feed-forward {\em output} component.

{\textbf{Multi-head Self-Attention Component.}  Here we denote the outputs of this component  at the \emph{l-th} computation block as $\mathbf{R}_Q^l$ and ${\mathbf{R}}_P^l$,  which are the generated vector representations   of the input question and passage respectively. And they are computed by the following Equation ~\ref{eq:sel}.}
\begin{equation}
	{\mathbf{R}_Q^l = MultiHeadSelfAttention(\mathbf{H}_Q);\qquad \mathbf{R}_P^l = MultiHeadSelfAttention(\mathbf{H}_P)\label{eq:sel}} 
\end{equation}

{In above equation, the computation method for the \emph{multi-head self-attention} operation is the same as the one in  \emph{Transformer}'s decoder \cite{vaswani2017attention}, thus readers can read the original paper of \emph{Transformer} for more detailed information if necessary.}

\begin{figure}[t]
	\centering  
	\includegraphics[width=0.70\linewidth]{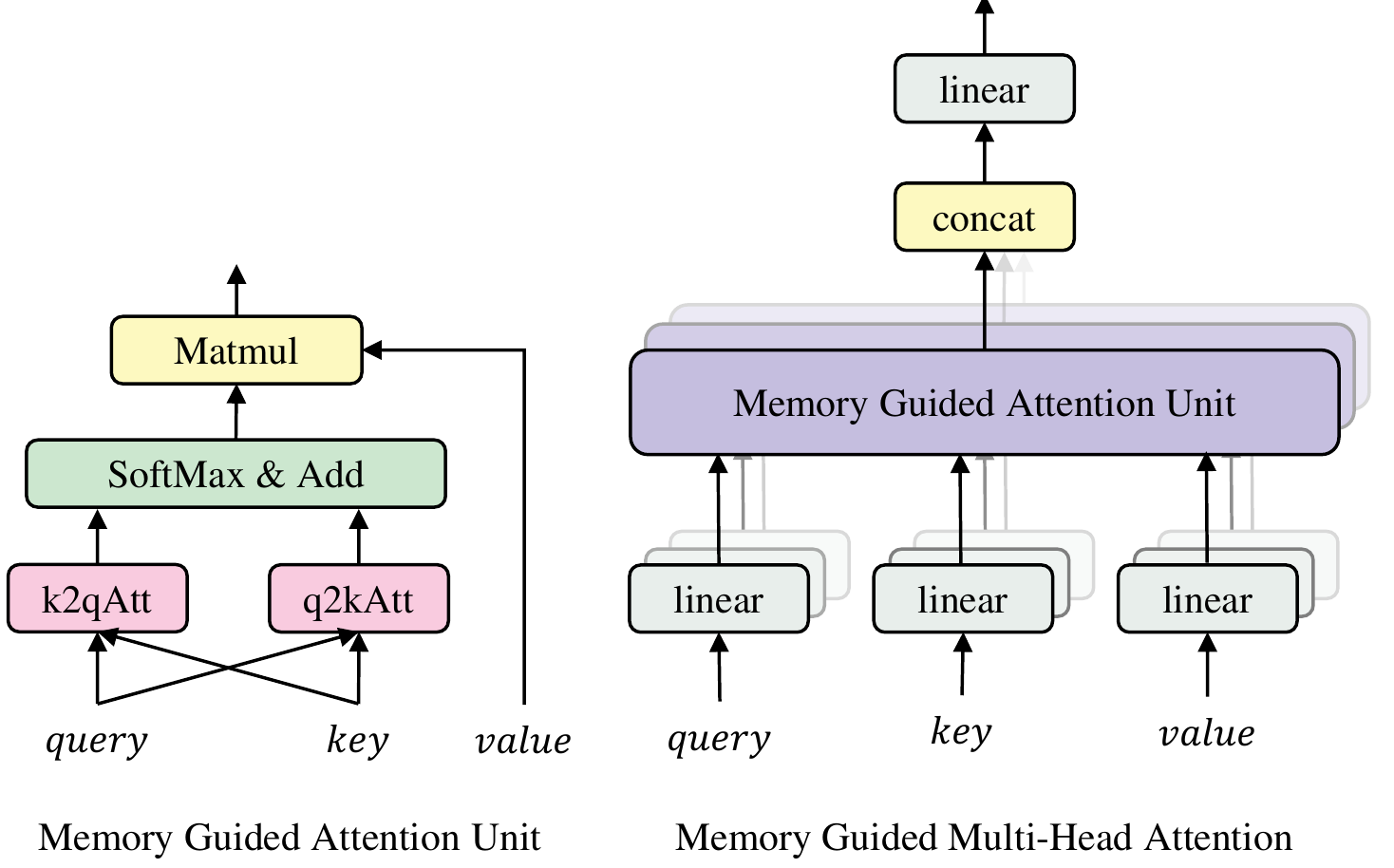}
	\caption{Illustration of the \emph{Memory-Guided Multi-Head Attention} method. }
	\label{fig:similarity}
\end{figure}

\noindent\textbf{Memory-Guided Interaction Mining Component.} Here we design a \emph{memory-guided multi-head} attention method to  mine the interactions between the input question and passage.  Its structure is shown in the right part of   Fig. \ref{fig:similarity} where ``\emph{query}" and  ``\emph{key}" denote the representations of question and passage respectively. During the interaction mining process, both ``\emph{query}" and  ``\emph{key}"  are dynamically updated guided by   ``\emph{value}". ``\emph{value}"  stores some ``\emph{memory}" information about the input question and passage, and would keep unchanged during the interaction mining process.

The basic component in this proposed interaction mining method is a ``\emph{memory-guided  attention unit}" whose structure is shown in the left part of  Fig. \ref{fig:similarity}. We can see that this component first computes two kinds of attentions: one is a \emph{key-to-query} attention (denoted as \emph{k2qAtt} in  Fig. \ref{fig:similarity}), and the other is  a \emph{query-to-key} attention (denoted as \emph{q2kAtt} in  Fig. \ref{fig:similarity}). Then these two kinds of attentions are integrated with the ``\emph{value}" part. 

Formally, we use use $\mathbf{Q}$, $\mathbf{K}$, and $\mathbf{V}$ to denote the  vector representations of ``\emph{query}", ``\emph{key}", and ``\emph{value}"    respectively . Then  the ``\emph{memory-guided  attention unit}" computes the attention  with Equation \ref{eq:mh00}, where $d$ is the scaling factor as the one defined in \emph{Transformer}.   
\begin{equation}
\begin{aligned}
{\rm MemAtt}\left(\mathbf{Q}, \mathbf{K}, \mathbf{V}\right) = [softmax\left(\frac{\mathbf{Q}\mathbf{K}^\top}{\sqrt{d}} \right)+ softmax\left(\frac{\mathbf{K}\mathbf{Q}^\top}{\sqrt{d}} \right)]\mathbf{V} \label{eq:mh00}
\end{aligned}
\end{equation}

We denote $\mathbf{H} = \mathbf{H_P} + \mathbf{H_Q}$ (here ``$+$" is an element-wise add operation, note as described in the  section \ref{sec:endoder}, $\mathbf{H_P}$ and $\mathbf{H_Q}$ have been padded into the same dimension). Based on $\mathbf{H}$, we define  ``\emph{value}" as $\mathbf{HW^V}$.   Accordingly, the ``\emph{memory-guided multi-head attention}" at the \emph{l-th} computation block can be written with following Equation \ref{eq:mh} and \ref{eq:mh1}, where $h$ is the number of the heads (we set $h$ to 8 in experiments),  $\mathbf{W}_i^Q$, $\mathbf{W}_i^K$,  $\mathbf{W}_i^V$,  $\mathbf{W}^O$  are trainable parameter matrices for the $i$-th head, and \emph{ReLU} is the activation function that is widely used in diverse neural models. 
\begin{equation}
\mathbf {O_{mrc}^l} = ReLU(Concat ({\rm head_1^l}, ..., {\rm head_h^l}  ) \mathbf{W}^O )\label{eq:mh} 
\end{equation}
\begin{equation}
{\rm head_i^l} = {\rm MemAtt}\left(\mathbf{R}_Q^l\mathbf{W}_i^Q,{\mathbf{R}}_P^l\mathbf{W}_i^K,\mathbf{H}\mathbf{W}^V\right)\label{eq:mh1} 
\end{equation}

%Then as shown in the right part of Figure \ref{fig:similarity}, the generated attention results are integrated into an multi-head attention framework, as shown in Following Equation \ref{eq:mh}.

%Finally, we can generate a  question-aware representation for the input passage by a  normalization operation with Equation \ref{eq:mh0}.
%\begin{equation}
%\mathbf{R}_P^l = {\rm Norm}\left(MultiHead\left(\mathbf{Q}, \mathbf{K}, \mathbf{V}\right)^l + {\mathbf{R}}_P^l\right) \label{eq:mh0} 
%\end{equation}

From above process we can see that  {during training,  as the computation blocks iterated, both   $\mathbf{Q}$ and  $\mathbf{K}$ are repeatedly updated, while  $\mathbf{V}$ keeps unchanged and will be reloaded at each iteration.} 

{Compared with the multi-head attention  method used in \emph{Transformer}~\cite{vaswani2017attention}, there are two important improvements in our method. 
First, it computes a kind of cross attentions that mine interactions from both the directions of ``\emph{query to key}'' and ``\emph{key to query}'', which can mine more richful and comprehensive interactions between questions and passages. This process is somewhat like a human's reading pattern that  understand the semantic meanings of questions by taking passages as context, and vice versa. So it would be  much helpful for \emph{understanding} the semantic meanings of questions and passages. Second,  its whole process is guided by a memory cell $\mathbf{H}$  where the original information of the question and passage is stored. This is very necessary for an MRC task: without $\mathbf{H}$, the mined interactions would ``\emph{forget}" more and more original input information as the  computation blocks iterated. However, the original information is the foundation of interactions, so ``\emph{forget}" them will increase the risk that the mined interactions are actually irrelevant to what are needed. In fact, setting a memory cell to store original input information is in line with a human's reading pattern that keeping the input text in mind when finding the answer. 
So our interaction mining method is  superior to existing  methods like BiDAF~\cite{seo2016bidirectional} or the original decoder in \emph{Transformer}~\cite{vaswani2017attention} by nature since lots of existing researches have proven that imitating human's reading patterns in an MRC model always brings  performance gains (see previous \emph{Related Work} section for details).} %In other words, the interaction mining process of our method is somewhat like human's can effectively address the risk of ``\emph{input information vanishing}" that is very harmful for an MRC task. %: the needed interactions should be highly related to both the input question and passage. 

\noindent\textbf{Output Component.} 
As shown in  Fig. \ref{fig:frame} (c),  the output of the last computation block in the \emph{memory guided interactions mining} component will be fed into a feed forward network in each computation block. We denote this output   as $\mathbf {O_{mrc}^N}$. Then  we perform two \emph{softmax} based operations to predict the probabilities of each token in the passage being the start and end positions of an answer  with following Equation \ref{eq:rst}, where $\mathbf{W}_s, \mathbf{W}_e $, $b_s$, and $b_e$ are learnable parameters.  
\begin{equation}
\begin{aligned}
&P_s = softmax(\mathbf{W_s O_{mrc}^N} + b_s); \\
&P_e=softmax(\mathbf{W_e O_{mrc}^N} + b_e)
\label{eq:rst}
\end{aligned}
\end{equation}

Finally,  following loss function is  defined to train the MRC module, where   $a^{s}_i$ and $a^{e}_i$ denote the start and end positions for the answer of the $i$-th sample. 
\begin{equation}
\mathcal{L}_{MRC}= - \frac{1}{2M}\sum_i^M{ [ log (P_s^{a_i^s})  + log (P_e^{a_i^e})]}
\end{equation}

%\begin{equation}
%\mathcal{L}_{\beta}\left(\theta_S, \theta_N\right)=-\frac{1}{2M}\sum_{i=1}^M \left[\log P^{begin}\left(a^{begin}_i\right) + \log P^{end}\left(a^{end}_i\right)\right]
%\end{equation}

 %,   ${\theta}_{M}$ is the private parameter set in the  MRC,  and ${\theta}_{S}$ is the parameter set shared by {NLI} and MRC.

\subsection{NLI Module}
{As analyzed above, we convert the traditional NLI task into the task of judge whether two question have the same semantic meaning. Accordingly, it is a natural way to  design a classification based NLI module here. Specifically, the training set  for the converted NLI task can be denoted as  $\mathcal{D}^{N}=\{(s_1^i, s_2^i, s^i)\}_{i=1}^N$ where $N$ is the number of samples.  In this set, each sample consists of following three items: the first sentence $s_1$, the second sentence $s_2$, and the answer $s$ ($s \in \{0, 1\}$) whose value indicates whether these two sentences have the same semantics meaning or not.}

{In BERT (or other pre-trained language models like RoBERTa) based models, the embedding representation of the padding token \emph{CLS} is believed to contain the global information of the whole input text. Thus, here $\boldsymbol{h}_{<CLS>}$ (the context-aware vector representation of  {\emph{ <CLS>}, see the descriptions in the \ref{sec:endoder} section}) is  used as a  contextualized sentence-level  representations of the input two questions.} Taking $\boldsymbol{h}_{<CLS>}$  as input, the NLI module uses following affine function to score the probability of the two input sentences having the same semantic meaning. 
\begin{equation}
\hat{y} = softmax\left(\mathbf{w}\boldsymbol{h}_{<CLS>} + b\right)
\label{eq:nli1}
\end{equation}
where $\mathbf{w}$ and $b$ are learnable parameter.

Finally,  following  loss function is  defined to train the NLI module, where $y_i \in \{0, 1\}$ denotes the true label for the $i$-th sample. 
\begin{equation}
\begin{aligned}
\mathcal{L}_{NLI} =  -\frac{1}{N}\sum_{i=1}^N[y_i\log\hat{y}_i + (1 - y_i)\log(1-\hat{y}_i)]
\end{aligned}
\end{equation}

 %,  ${\theta}_{N}$ is the private parameter set in {NLI}, and ${\theta}_{S}$ is the parameter set shared by {NLI} and MRC.

\subsection{Multi-Task and Multi-Language Learning}
We use a   multi-task learning framework to train the modules of MRC and {NLI}  simultaneously.  The whole loss function is defined with Equation \ref{eq:total_loss}.    
\begin{equation}
\mathcal{L}=\alpha  \mathcal{L}_{NLI}({\theta}_{N},{\theta}_{S}) + \beta\mathcal{L}_{MRC}({\theta}_{M},{\theta}_{S})
\label{eq:total_loss}
\end{equation}

where $\alpha$ and $\beta$ are two hyperparameters which are set to 0.5 and 1 respectively in our experiments,   $\theta_S$ represents the task-independent parameter set shared by  the modules of MRC and {NLI},   $\theta_M$ and $\theta_N$ are the  task-dependent parameter sets for the MRC module and the {NLI} module respectively.
 %In experiments, we .

\begin{algorithm}[t]
	\caption{ Algorithm of the Multi-Task and Multi-Language Learning.}
	\label{alg:mtl}
	\begin{algorithmic}[1]
		\Require
		the NLI dataset $\mathcal{D}^{N}$ and the mixed MRC dataset $\mathcal{D}^{M}$ =\{ $\mathcal{D}^{M}_{{lg}_{1}}$, $\mathcal{D}^{M}_{{lg}_{2}}$, ..., $\mathcal{D}^{M}_{{lg}_k}$ \} where each sub-dataset corresponds to an MRC dataset  of a specific language.\
		\Ensure
		\State Preprocessing: compute a Bernoulli distribution based sampling probability  for each sub-dataset in the mixed MRC dataset according to the ratio of the number of samples in this sub-dataset to the total number of samples in the mixed MRC dataset.
		\State Sampling: randomly select which task to be trained. 
		\Statex \quad  If  NLI is selected, randomly select a 		sample  from $\mathcal{D}^{N}$. 
		\Statex \quad  Else,  select a sample from an MRC sub-dataset  according to its sampling probability.
		\State Generate a batch: repeat above sampling step until selecting a predefined number of samples to build up a batch $B$.
		\label{code:fram:sample}
		\State Training: joint train the MRC and NLI modules on $B$ with the multi-task learning based method. 
		\State Iteration: repeat above steps of 2-4 until reaching the predefined training epoch.
		%		\end{algorithmic}
	\label{code:fram:trainbase}
\end{algorithmic}
\end{algorithm}

Besides, as analyzed above that an MRC model tends to perform well on in-domain test sets but perform poorly on out-of-domain test sets, which is mainly caused by the similar distributions between  training sets and test sets. To overcome this issue, we generate a mixed MRC training set which consists of several MRC training sets of different languages.  Then the MRC module is trained based on the samples in this mixed training set.  For example, one can form a mixed training set by combining an English MRC training set with a Chinese MRC training set, but still testing the MRC model on an English test set (or a Chinese test set). By this way, the distributions of the training set and the test set will be completely different,  which will push the MRC model learn more generalized knowledge during training because none of a specific language's training set could dominant the process of  model training. Accordingly, the generalization issue will be alleviated greatly. We call this new  training method as  \emph{multi-language learning},  it is then combined with the multi-task learning method to form a new \emph{multi-task and multi-language learning} mechanism. 

\begin{figure*}[t]
	\centering
	\includegraphics[width=0.97\linewidth]{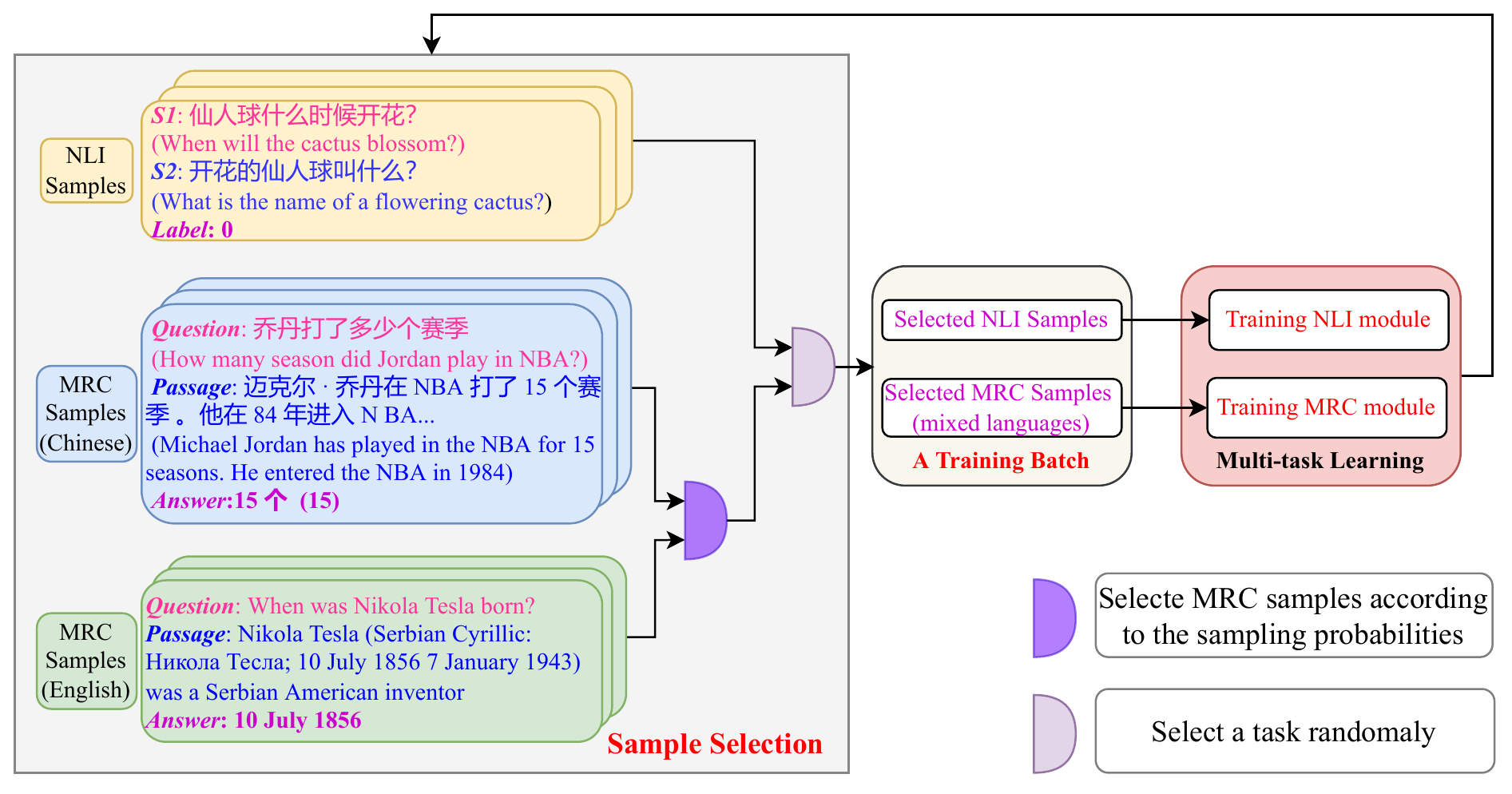}
	\caption{{Illustration of the \emph{multi-task and multi-language learning}.}}\label{fig:mtl}
\end{figure*}

{Specifically, the detailed process of this new learning mechanism is shown in  Algorithm \ref{alg:mtl}. The input of this algorithm includes an NLI dataset and a mixed MRC dataset. In the \emph{multi-task learning} mechanism, the samples of different tasks are randomly selected. However, for the mixed MRC dataset, if its sub-datasets of different languages  have the same probability of being used as a data source to select training samples, the samples in the smaller sub-datasets would be over-trained while the samples in the larger sub-datasets would be under-trained. To overcome this problem, we assign different selected probabilities for the sub-datasets of different languages.  Our basic idea is that the larger the number of  a sub-dataset, the more possible its samples should be selected. In our algorithm, we design a Bernoulli distribution based  method to compute a \emph{sampling probability} for each sub-dataset in the mixed MRC dataset, as shown in the \emph{Preprocessing} step in our algorithm. Based on these probabilities, the samples in different sub-datasets in the mixed MRC dataset are selected and batched with the samples in the NLI dataset, then the batched samples are jointly trained with the \emph{multi-task learning} mechanism. To more clearly demonstrate our Algorithm \ref{alg:mtl}, we further use Fig. \ref{fig:mtl} to illustrate its  whole learning process with some concrete samples.}

{It should be noted that our \emph{multi-language learning} does not focus on the effort of making a model ``\emph{see}'' some samples that are expected to be present during testing. Thus it is much different from existing data augmentation based methods.}  {Beside, our \emph{multi-language learning} is also much different from some existing models like the one that uses \emph{multi-lingual pre-trained language models} ~\cite{hsu-etal-2019-zero} or the one that uses \emph{cross-lingual pre-trained language models}~\cite{2021From}. Although all of  these existing models are trained on datasets of a source language but tested on datasets of a target language, all of them heavily depend on the pre-trained \emph{multi-lingual} or \emph{cross-lingual}   language models. In contrast, our method does not use any of these kinds of language models. During training, the tokens of different languages learn their own embedding representations, while the model parameters are shared among samples of different languages.   If a token could not be found in the vocabulary of a pre-trained language model, it would be regarded as an out-of-vocabulary token and its embedding would be randomly initialized and updated during model training.} 
{As the study of ~\cite{hsu-etal-2019-zero} shows that tokens from different languages might be embedded into the same space with close spatial distribution. Their study further shows that even though during the fine-tuning only data of a specific language is used, the embedding of tokens in another language changed accordingly. These results show that the \emph{multi-language learning} does have the capability of pushing the model learn more generalized knowledge during training, which is much helpful for alleviating the mentioned \emph{generalization} issue.}%Generalization issue which refers to models usually perform well on in-domain test sets yet perform poorly on out-of-domain test sets.
%Figure \ref{fig:mtl} illustrates   the process of \emph{multi-task and multi-language learning}  with a concrete example.

\section{Experiments}

\subsection{Experimental Settings}
\textbf{MRC Datasets} In this study, we evaluate the performance of our model on the following benchmark MRC datasets, all of which are designed to  measure the robustness of an MRC model, including the abilities of addressing the issues of over sensitivity, over stability and generalization.

\textbf{(i) DuReader (robust).} DuReader (robust) \cite{tang-etal-2021-dureader} is  a large benchmark Chinese MRC dataset released in the MRC Competition of ``\emph{2020 Language and Intelligence Challenge}\footnote{\url{https://aistudio.baidu.com/aistudio/competition/detail/28}}" (LIC-2020). It is a variant  of DuReader \cite{he2018dureader} and  is  designed to measure an MRC model's ability for addressing the issues of over sensitivity, over stability and generalization.  %It takes EM and F1  as evaluation metrics. 
%The dataset includes training set, development set and test set.
Its training and development sets  consist of 15K and 1.4k samples respectively. In these two sets, the questions  are all  real questions issued by users in \emph{Baidu} search engine. The passages are extracted from the search results of \emph{Baidu} search engine and \emph{Baidu Zhidao} (a question answering community). 

The test set of DuReader (robust) includes four subsets. (i) \emph{In-domain subset}:  the construction method and source of this subset is the same as the training set and the development set. (ii) \emph{Over sensitivity subset}: the samples in this subset are randomly sampled from the \emph{in-domain} subset. The new samples are constructed by paraphrasing the questions.  But the questions generated here are all real questions asked by users in \emph{Baidu} search engine.  (iii) \emph{Over stability subset}: the samples in this subset are sampled from the \emph{in-domain} subset by rules, but annotated by human experts. The passages in this subset are all real passages.  (vi) \emph{Generalization subset}: the samples in this subset have a different distribution from that of the training set. The samples include real questions and passages extracted from \emph{educational} and \emph{financial} passages. There are 1.3K, 1.3K, 0.8K, and 1.6K samples in these four subsets respectively.

DuReader (robust) divides its test set into two versions. The first one  consists of  most of samples in the in-domain subset  and a small portion of samples in the other three robustness subsets. And the second one  consists of all the four subsets. For convenience, we denote these two versions' test sets as {Test1} and {Test2} respectively.
It should be noted that DuReader (robust) DOES NOT release its answer labeled test set  to avoid models achieve  overestimated performance by a test set guided  training.  Instead,  researchers have to submit their results to the competition organizers so that the performance of their models can be evaluated. % making some elaborate training strategies  
%\noindent\textbf{SQuAD1.0} For the \emph{multi-language learning}, we use SQuAD1.0 \cite{rajpurkar2018know} as an extra MRC dataset. SQuAD is a widely used MRC benchmark dataset  consisting of 100,000+ questions and the answer to each question is a segment of text from the corresponding reading passage. 

\textbf{(ii) SQuAD-related Datasets.} To explore the robustness of MRC models, \cite{gan2019improving} create two {\bf TEST} sets consisting of paraphrased questions that  are generated by taking some questions in the development set of SQuAD 1.0 ~\cite{rajpurkar2018know} as source questions. The first test set is a non-adversarial paraphrased set that is generated with a neural paraphrasing model trained on a dataset where each sample has a tuple form of (\emph{source question, multiple paraphrase suggestions}). The paraphrased results in this test set are subsequently verified by human annotators. The second one is an adversarial paraphrased test set that is generated manually by going through question and context pairs from the SQuAD development set and re-writing the  question using context words  near a confusing answer candidate if such a candidate exists and there are suitable nearby context words for use in paraphrasing. 
We denote these two test sets as SQuAD (Non-Adv-Paraphrased) and SQuAD (Adv-Paraphrased) respectively.  The first test set   consists of 1,062 questions and the second test set consists of  56  questions. % paraphrased using context words near a confusing answer candidate respectively. %These two test sets are designed to test MRC models’ over sensitivity and over stability respectively. % We use these two test sets to evaluate the robustness of our model under an extreme condition: trained with Chinese data but tested with  Engligh data.  

\noindent\textbf{NLI Dataset} Here LCQMC\footnote{\url{http://icrc.hitsz.edu.cn/info/1037/1146.htm}} \cite{liu2018lcqmc}  is used as the training set for  the {NLI} module. LCQMC is a dataset that  focuses on the intent matching of two sentences. Thus it is suitable to  be used to train our NLI module.  Totally, this dataset contains 260,068 question pairs with manual annotations. And it is divided into three parts: a training set that contains  238,766 question pairs, a development set that contains 8,802 question pairs, and a test set that contains 12,500 question pairs. %contain 238,766 question pairs, the development set contains 8,802 question pairs, and the test set contains 12,500 question pairs.

\noindent\textbf{Multi-language Learning Setting} In the multi-language learning, we use  SQuAD 1.0 as an auxiliary MRC dataset. SQuAD 1.0 is a widely used large scale English MRC benchmark dataset.  Here we select it mainly due to the following two reasons. First,  both SQuAD 1.0 and DuReader (robust) belong to the single-passage MRC dataset\footnote{There is a kind of multi-passage (also called multi-document) MRC datasets that provides multiple passages (or documents) for each question, like DuReader, MS MARCO, TriviaQA (web)~\cite{joshi2017triviaqa}, etc.}. Second,  the average lengths of passages in both datasets are close. These two characteristic of SQuAD 1.0 allow us to  concentrate on the robustness issues other than some data preprocessing work.   Besides,  there are  more than 100,000 questions in SQuAD 1.0, which will make the distributions of the training set and the test set different greatly. Thus it helps to provide an ideal platform to  evaluate the generalization of an MRC model. % consists of and the answer to each question is a segment of text in  the corresponding  passage

%Here we do not  use other datasets or use more datasets, which  is mainly due to following reasons. First, both SQuAD1.0 and DuReader (rubost) is designed for the single-passage MRC task. Second,  the average lengths of passages in both datasets are close.    Third, \cite{gan2019improving} point out that the simpler  datasets allow researchers to concentrate on the robustness of MRC models.

%which is a large benchmark English MRC dataset and the 
%\noindent\textbf{LCQMC} We use LCQMC\footnote{\url{http://icrc.hitsz.edu.cn/info/1037/1146.htm}} \cite{liu2018lcqmc} as the dataset for the \emph{NLI} task.  LCQMC is more general than paraphrase corpus as it focuses on intent matching rather than paraphrase. The corpus contains 260,068 question pairs with manual annotation. Its training set contains 238,766 question pairs, the development set contains 8,802 question pairs, and the test set contains 12,500 question pairs. 

\noindent\textbf{Other  Settings} A Chinese {RoBERTa}    \cite{cui2019pre} is used as the required \emph{shared encoder module}. In subsequent sections,  for all the mentioned   language models, we use their large versions. The ensemble model  is obtained by averaging 4 single models’ prediction probabilities. 
Extract match (EM) and F1 are used as evaluation metrics.   During training,  AdamW~\cite{KingmaB14}  is used to train our model and word embeddings are not updated. Based on the results on development sets, the learning rate, batch size, and  training epoch are set to 0.001, 16,   and 3 respectively.  % is set to   3.  

\begin{table}[t]
	\centering
		\caption{Main Results on DuReader (robust).}
\label{tab:main}%

	{\begin{tabular}{lcc}
			\toprule
			\multicolumn{3}{c}{Test1} \\
			\hline
			\multicolumn{1}{l}{Model} & F1    & EM \\
			\hline
			BERT (zh) \cite{devlin2018bert}  & 70.74 & 54.45 \\
			XLNet (zh) \cite{yang2019xlnet} & 68.79 & 53.70 \\
			RoBERTa (zh) \cite{cui2019pre} & 73.76 & 56.75 \\
			\hline
			\multicolumn{1}{l}{Our Model (Single)} & \textbf{80.39} & \textbf{66.55} \\
			\multicolumn{1}{l}{Our Model (Ensemble)} & \textbf{82.70} & \textbf{68.50} \\
			\midrule
			\multicolumn{3}{c}{Test2} \\
			\hline
			Our Model (Ensemble) & \textbf{79.45} & \textbf{64.76} \\
			\hline
		\end{tabular}
	}%
	{
		\caption{Ablation Experiments on Test1 of DuReader (robust).  ``- \emph{Memory-Guided Multi-head Attention}"  means replacing it with a common multi-head attention method as used in \emph{Transformer}'s decoder. ``\emph{zh}'' denotes the corresponding model is a Chinese version. }
\label{tab:abl}		
		\begin{tabular}{lcc}
			\toprule
			\multicolumn{1}{l}{Model} & F1    & EM \\
			\hline
			Our Model (Single) & \textbf{80.39} & \textbf{66.55} \\
			\hline
			- \emph{NLI} module & 76.26 & 60.6 \\
			- \emph{Multi-language learning}   & 77.47 & 62.8\\
			- \emph{Memory-Guided Multi-head Attention}  & 78.65 & 64.5 \\
			\hline
		\end{tabular}%
}
\end{table}%

\subsection{Experiments on DuReader (robust)}

%DuReader (robust) divides its test set into two parts: the first one  (we denote it as \emph{Test1}) contains most of the in-domain test set and a small portion of the robustness sets, the second one ( we deonte is as \emph{Test2}, its test set leader board  is closed now) contains all the four test sets. % on which the final performance is offically evaluated. 

\noindent\textbf{Main Results} The main results on DuReader (robust) are shown in Table \ref{tab:main}. On Test1, we can see that our model achieves much  better results than all the compared strong baselines. Especially, our model achieves much better results than RoBERTa  which has  achieved much better results than many existing Chinese pre-trained models (like BERT  and ERNIE \cite{sun2019ernie}) on various natural language processing tasks including several Chinese MRC tasks \cite{cui2019pre}. These results show that our model is very effective and it can better address the robustness issues. 

In fact, we participated in the  MRC Competition of LIC-2020. Finally, our model ranked \textbf{NO.2} on the test set leader board of Test1. On Test2,  our model  achieved very competitive results again: it ranked \textbf{No.3}  on this full DuReader (robust) test set leader board\footnote{{Here we could not compare our model with the  top 2 models, because we could not find any papers about their model details either in conferences, journals, or on arXiv.}}. 

Note that there are no detailed comparison results on Test2. This is because the final competition results of LIC-2020 were decided by the results on Test2. And LIC-2020 took  a closed way to evaluate an MRC model's performance: researchers must submit their results within a specified deadline, and there was a submission limit per system per day. Thus, researchers usually tried to find the best model based on the results on Test1, and then tried to obtain the best   results on Test2 by tuning the selected model. In other words, there was almost no chance for researchers to compare the performance of different models on Test2. 

Now, the competition of LIC-2020 was closed, thus we could not make more detailed comparisons with the latest state-of-the-art MRC models. And we leave such comparisons in  subsequent sections. %have to   limit for cannot see their models' evaluation results in real time.

%We can see that our model achieves much  better results than all the compared state-of-the-art baselines. In particular, all the baselines  outperform human on many MRC benchmark datasets like SQuAD.  In fact, our model ranks No2. and No.3 on the leader board of Test1 and Test2 respectively. These results show that our model is very robust.

\noindent\textbf{Ablation Results} To demonstrate the contributions of different modules in our model, we conduct ablation experiments on Test1 and the results are shown in Table \ref{tab:abl}. 

From these results we can draw following conclusions. First, each component of our model is helpful for improving the performance.  Second, {NLI} plays more  roles than the other two modules, which indicates that the key of developing a robust MRC model is to precisely \emph{understand} the semantic meanings of input questions. Third, introducing a completely different language's MRC training set is helpful for improving the robustness of an MRC model. Fourth, the proposed \emph{memory-guided multi-head attention} is effective for addressing the robustness issues and it performs better  than the traditional   \emph{multi-head attention} method.

\begin{figure}[t]
	\centering
	\includegraphics[width=0.95\linewidth]{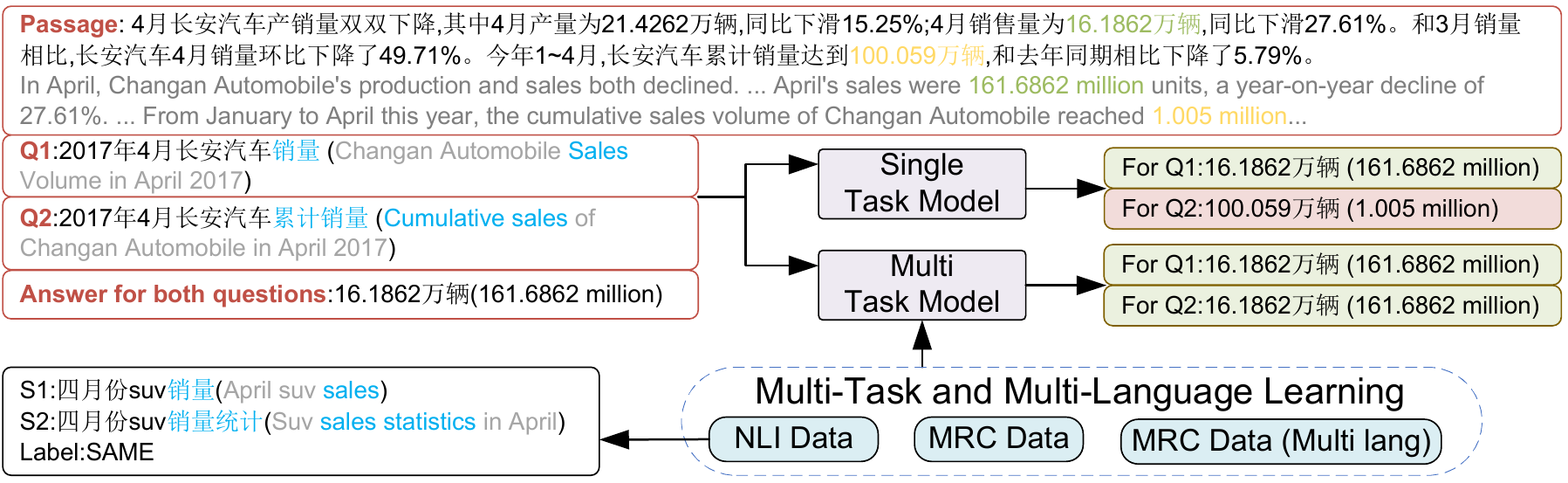}
	\caption{A Case Study (The words with the same color have the same meaning. ).}\label{fig:cs}
\end{figure}

\noindent\textbf{Case Study}  Fig. \ref{fig:cs} illustrates a case study of our model. In this example, the semantic meanings of two questions (Q1 is the source question and Q2 is the paraphrased question) are the same. We can see that if we do not use the NLI module, the model (``\emph{Single Task Model}'' in  Fig. \ref{fig:cs}) outputs a wrong answer for the paraphrased question (Q2). In contrast, the full model precisely distinguishes the semantic meanings of these two questions and  outputs correct answers for both of them.  These results further confirm the effectiveness of the proposed model for addressing the robustness issues. % in MRC. 
%The main results of our model are shown in Table \ref{tab:main}. We can see that our model achieves better results than all the compared state-of-the-art baselines. In particular, all the baselines  outperform human on many MRC benchmark datasets like SQuAD.  In fact, our model ranks No2. and No.3 on the leader board of Test1 and Test2 respectively. 

\subsection{Experiments on SQuAD-related Datasets}
On  SQuAD (Non-Adv-Paraphrased) and SQuAD (Adv-Paraphrased), we  evaluate the robustness of our model based on following two kinds of extreme  evaluations: (i)  testing our model that is designed for DuReader (robust)  on these two English test sets directly; and (ii) comparing an English version of our model with the models that are retrained by a data augmentation based method.   

Obviously, both kinds of evaluations are significant unfair to our model. For the first kind of evaluation,   although there is a multi-language learning mechanism, most of the modules in our model, including the shared encoder module and the NLI module,  are trained on Chinese datasets. For the second kind of evaluation,  our model would not use any additional training data while some compared baselines would be retrained by the data augmentation based method. However, we think both kinds of evaluations are very meaningful: the first kind of evaluation  provides an ideal scenario to evaluate  the generality ability of an MRC model because the distributions of training sets and test sets are completely different; and the second kind of evaluation provides an ideal scenario to evaluate the true potentiality of an MRC model for addressing the robustness issues because the data augmentation based methods are not always available, especially in some cases where constructing the augmented data is time-consuming and high-cost.   %than the because  . %And for the second one,  it can further indicate the generalization of an MRC  model.

In subsequent experiments, we use the single version of our model for evaluations, and all the  models marked by $^\dagger$ denote their corresponding  results  are directly copied from \cite{gan2019improving}.
%However, we think such evaluation is very meaningful because 

\noindent\textbf{Extreme  Evaluation 1.} 
%In this kind of extreme evaluation, we evaluate the performance the Chinese version of our model  directly on the two English test sets of  SQuAD (Non-Adv-Paraphrased) and SQuAD (Adv-Paraphrased). 
The  results of the first kind of extreme evaluation are shown in Table \ref{tab:extre} and \ref{tab:extre1}. We can see that on both test sets, our model achieves very competitive results. First, when compared with the models that are fully trained under the common settings,  our model achieves significant better results than DrQA and BiDAF. Our model also achieves close results to the well trained BERT on the robust parts (denoted as \emph{P-Questions} and \emph{A-Questions} in Table \ref{tab:extre} and \ref{tab:extre1}) of the two datasets. Second, the source questions (denoted as S-Questions in  Table \ref{tab:extre} and \ref{tab:extre1})  in    SQuAD (Non-Adv-Paraphrased) and SQuAD (Adv-Paraphrased) form two common MRC datasets, and  our model performs much well on them.  These results indicate that our model is effective on both the robustness datasets and the common datasets even under an extreme and unfair evaluation. % compared with both kinds of  baselines.

It should be noted that  \cite{hsu-etal-2019-zero} leverage the pre-trained multilingual BERT (multi-BERT) in cross-lingual zero-shot reading comprehension. That is,  multi-BERT is fine-tuned on data of a language, but is tested on data of another language. Obviously, this zero-shot setting is similar to the setting of our first kind of extreme evaluation. But here we do not take  multi-BERT as a baseline. This is mainly because we could not provide a fair platform for comparisons: if we fine-tune  multi-BERT on DuReader (robust) but test it on the mentioned SQuAD-related datasets, it would be unfair for  multi-BERT  because our model uses the training set of SQuAD  in the multi-language learning while multi-BERT not. This is also the reason why we do not compare our model with other state-of-the-art MRC models under this kind of extreme evaluation. %Thus we only compare these two models under the second kind of extreme evaluation where the comparison is fair because both models use the SQuAD (training) data and some multi-lingual resources.  
%These results show that our model is very robust and it has stable performance on both robust datasets and commone datasets. %, which is much important for real deployed. %because it indicates our model would also   %competitive results again.  

\begin{table*}[t]
	\centering
	%	\makebox[\linewidth]
	{
				\caption{Extreme Evaluations 1 (a): on SQuAD (Non-Adv-Paraphrased). S-Questions and P-Questions refer to the source questions from  SQuAD's development set and		their corresponding paraphrased questions respectively. }
		\label{tab:extre}%
		
		\begin{tabular}{lllll}
			\hline
			\multirow{2}[4]{*}{Model} & \multicolumn{2}{c}{EM} & \multicolumn{2}{c}{F1}  \\
			%		\multicolumn{cols}{pos}{text}
			\cmidrule{2-5}          & \multicolumn{1}{l}{S-Questions} & \multicolumn{1}{l}{P-Questions} & \multicolumn{1}{l}{S-Questions} & \multicolumn{1}{l}{P-Questions}  \\
			\toprule
			\multicolumn{1}{l}{BiDAF$^\dagger$ {\cite{seo2016bidirectional}}} & 67.8  & 63.84 & 76.85 & 73.51\\
			\multicolumn{1}{l}{DrQA$^\dagger$ {\cite{chen2017smarnet}}} & 67.33 & 65.25 & 76.25 & 74.25\\
			\multicolumn{1}{l}{BERT$^\dagger$ {\cite{devlin2018bert}}} &{83.62} & {79.85} & {90.78} & {87.63}  \\
%			\hline
			
%			ALBERT (zh)~\cite{lan2020albert}&&&&\\ %&89.83&87.19&95.47&93.21\\ 
%			Retro-Reader (zh) \cite{zhang2021retrospective}&&&&\\% &88.42&85.78&94.06&92.19\\
			\hline
%			\multicolumn{1}{l}{RoBERta (zh)}  & 75.7  & 73.44 & 84.59 & 82.18  \\	
%				{  + NLI} & &&& \\
%				{  + NLI + MultLag}& &&& \\		
			{Our Model}& 77.91 & 76.90  & 87.38 & 85.87  \\
			\hline
		\end{tabular}

	}
	{		\caption{Extreme Evaluations 1 (b): on SQuAD (Adv-Paraphrased).  S-Questions and A-Questions refer to the source questions from  SQuAD's development set and		their corresponding adversarial paraphrased questions respectively.}
		\label{tab:extre1}%
		\begin{tabular}{lllll}
			\hline
			\multirow{2}[4]{*}{Model} & \multicolumn{2}{c}{EM} & \multicolumn{2}{c}{F1} \\
			%		\multicolumn{cols}{pos}{text}
			\cmidrule{2-5}         & \multicolumn{1}{l}{S-Questions} & \multicolumn{1}{l}{A-Questions} & \multicolumn{1}{l}{S-Questions} & \multicolumn{1}{l}{A-Questions} \\
			\toprule
			\multicolumn{1}{l}{BiDAF$^\dagger$ {\cite{seo2016bidirectional}}}  & 75    & 30.36 & 81.55 & 38.3 \\
			\multicolumn{1}{l}{DrQA$^\dagger$ {\cite{chen2017smarnet}}}   & 71.43 & 39.29 & 81.02 & 48.94 \\
			\multicolumn{1}{l}{BERT$^\dagger$ {\cite{devlin2018bert}}} &  82.14 & 57.14 & 89.31 & 63.18 \\
			\hline
			
%			ALBERT (zh)~\cite{lan2020albert}&&&&\\%&92.86&73.21&96.77&80.23\\
%			Retro-Reader (zh) \cite{zhang2021retrospective}&&&&\\%&89.29&76.69&93.%20&83.20\\
			\hline
%			\multicolumn{1}{l}{RoBERta(zh) {\cite{cui2019pre}}}   & 73.21 & 48.21 & 79.19 & 54.29 \\
%			{  + NLI} & &&& \\
%			{  + NLI + MultLag}& &&& \\		
			{Our Model}  & 77.36 & 56.79 & 85.74 & 62.22 \\
			\hline
		\end{tabular}

	}
\end{table*}%

\noindent\textbf{Extreme  Evaluation 2.} Based on the evaluation results  on SQuAD (Non-Adv-Paraphrased) and SQuAD (Adv-Paraphrased),  \cite{gan2019improving} argue that the original training dataset does not contain sufficiently diverse phrased questions, which leads to the models not learning to respond correctly to various ways of asking the same question. 
%They further argue that  a natural way to improve the robustness of MRC models to question paraphrasing would be to expose them to more diverse question phrasing. 
They further argue that the  capabilities of models for addressing the robustness issues can be improved by   a \emph{data augmentation} based method. So they use the similar methods as used when creating the mentioned two test sets to generate two additional training sets: the first one contains  25,000 non-adversarial paraphrased questions, and the second one contains 25,000 adversarial paraphrased questions. For simplicity, we denote these two additional training sets as \emph{Non-Adv-Additional Data} and \emph{Adv-Additional Data}, and denote the original training set and development set of SQuAD as SQuAD (training) and SQuAD (dev). With these additional data, \cite{gan2019improving}  report following four kinds of comparison results to demonstrate the effectiveness of the data augmentation method. (i) On SQuAD (Non-Adv-Paraphrased), the  results of  models between trained with  SQuAD (training)  and   ``SQuAD (training) +  \emph{Non-Adv-Additional Data}''; (ii) On SQuAD (Adv-Paraphrased), the  results of  models  between trained with  SQuAD (training)  and   ``SQuAD (training) + \emph{Adv-Additional Data}'';  (iii) On SQuAD (dev), the  results of  models  between trained with  SQuAD (training)  and   ``SQuAD (training) + \emph{Non-Adv-Additional Data}''; (iv) On SQuAD (dev), the  results of  models  between trained with  SQuAD (training)  and   ``SQuAD (training) + \emph{Adv-Additional Data}''. % With Then these two additional trainign sets are used to retrain all three MRC models with  %the original training data of SQuAD. % and the additional 25,000 paraphrased questions. 
%They report following two kinds of before/after re-training results: (i) the performance differences on the non-adversarial paraphrased test set (SQuAD (Non-Adv-Paraphrased)) and on the original development set (SQuAD (dev)), and (ii) the performance differences on the adversarial paraphrased test set (SQuAD (Adv-Paraphrased)) and on the original development set (SQuAD (dev)). 
However, \cite{gan2019improving} do not release these two additional training sets.  So we could not compare the performance of our model trained with these additional training sets with their reported results directly.  

\begin{table*}[t]
	\centering
		\caption{Extreme  Evaluation 2 (a): performance of different models before and after re-training on SQuAD (Non-Adv-Paraphrased) (the left part) and on  SQuAD (Adv-Paraphrased) (the right part).  } 
	\label{tab:gen1}
	\resizebox{\textwidth}{!}
	{\begin{tabular}{lllll|llll}
			\hline
			\multirow{2}[4]{*}{Model} & \multicolumn{2}{c}{EM} & \multicolumn{2}{c|}{F1} & \multicolumn{2}{c}{EM} & \multicolumn{2}{c}{F1} \\
			%		\multicolumn{cols}{pos}{text}
			\cmidrule{2-9}          & \multicolumn{1}{l}{Before} & \multicolumn{1}{l}{After} &
			\multicolumn{1}{l}{Before} & \multicolumn{1}{l|}{After} &\multicolumn{1}{l}{Before} & \multicolumn{1}{l}{After} &\multicolumn{1}{l}{Before} & \multicolumn{1}{l}{After} \\
			\midrule
			\multicolumn{1}{l}{BERT$^\dagger$ {\cite{devlin2018bert}}} &{79.85} & {80.89}   & {87.63} & {88.62} &57.14 &69.64   &63.18 &73.85    \\
			\multicolumn{1}{l}{DrQA$^\dagger$ {\cite{chen2017smarnet}}}  & 65.25&67.33   &74.25 &75 & 39.29&41.07 &48.94&49.86   \\
			\multicolumn{1}{l}{BiDAF$^\dagger$ {\cite{seo2016bidirectional}}}  & 63.84 &66.2   &73.51 &75.94  & 30.36&39.24 &38.3&47.49  \\
			\hline
			%			{RoBERTa-large  {\cite{cui2019pre}}} &-&85.22&-&-&92.00&-\\
			ALBERT-large ~\cite{lan2020albert} &-&{\bf 83.24} &-&{\bf 89.85} &-&69.64 &-&74.68 \\ 
			Retro-Reader  \cite{zhang2021retrospective}&-&82.20& -&89.40 &-&69.64 &-&75.04  \\
			\hline
			%			\multicolumn{1}{l}{RoBERta(zh) {\cite{cui2019pre}}}   &&&&&& \\
			%			{  + NLI} & &&&&& \\
			%			{  + NLI + MultLag}& &&&&& \\		
			{Our Model} &76.90 &{82.96}   &85.87 &{89.78} &56.79&{\bf 76.79} &62.22&{\bf 80.92}\\
			\hline
	\end{tabular}}
	
\end{table*}%
\begin{table*}[t]
	\centering
	\caption{Extreme Evaluation 2 (b): performance of different models on SQuAD (dev)  before and after re-training. For the baselines marked by $^\dagger$, After$_1$ and After$_2$ denote the results  re-trained with additional \emph{Non-Adv-Additional Data} and  \emph{Adv-Additional Data} respectively. For other baselines and our model, we  use  After$_1$ to denote their English versions.} 
\label{tab:gen2}
	{\begin{tabular}{llll|lll}
			\hline
			\multirow{2}[4]{*}{Model} & \multicolumn{3}{c|}{EM} & \multicolumn{3}{c}{F1}  \\
			%		\multicolumn{cols}{pos}{text}
			\cmidrule{2-7}          & \multicolumn{1}{l}{Before} & \multicolumn{1}{l}{After$_1$}  &
			\multicolumn{1}{l|}{After$_2$} &
			\multicolumn{1}{l}{Before} & \multicolumn{1}{l}{After$_1$}  &
			\multicolumn{1}{l}{After$_2$}  \\
			\midrule
			\multicolumn{1}{l}{BERT$^\dagger$ {\cite{devlin2018bert}}}   & 84.02 & 83.76   &83.33 & 91 &90.88  &90.49 \\
			\multicolumn{1}{l}{DrQA$^\dagger$ {\cite{chen2017smarnet}}}   &69.04 &68.74  &67.93  &78.38 &77.86  &77.45 \\
			\multicolumn{1}{l}{BiDAF$^\dagger$ {\cite{seo2016bidirectional}}}  &67.67 &67.49  &66.23 &77.46 &77.1   &76.19  \\
			\hline
			%\multicolumn{1}{l}{RoBERta(eg) {\cite{cui2019pre}}} & &&&&  \\
			%			{RoBERTa-large  {\cite{cui2019pre}}} &-&85.91&-&-&-\\
			ALBERT-large ~\cite{lan2020albert} &-&84.56&-&-&91.63&- \\ 
			Retro-Reader  \cite{zhang2021retrospective}&-&84.09&-&-&91.09&- \\
			\hline
			%			\multicolumn{1}{l}{RoBERta(eg) {\cite{cui2019pre}}} & &&&&  \\
			%			{  + NLI}  &&&&& \\
			%			{  + NLI + MultLag} &&&&& \\		
			{Our Model}  &78.42 & {\bf 85.50}  &-& 87.35 & {\bf 92.46} & - \\
			\hline
	\end{tabular}}

\end{table*}%

Instead,  we make another kind of extreme evaluation by comparing the English version of our model with the models retrained by using the mentioned two additional training sets. To this end, we make two modifications on our model. First, we use the Quora Question-pair  dataset\footnote{\url{https://www.kaggle.com/c/quora-question-pairs/data?select=train.csv.zip}}  as the trainign set for the {NLI} module. Second,  we change  RoBERTa to its  English version. Besides, we also take several state-of-the-art models as baselines.  In experiments, to make a fair comparison, we replace the version of AlBERT used in   Retro-Reader~\cite{zhang2021retrospective} from \emph{xxlarge} to \emph{large} since as mentioned previously that for all the  language models,  we use their large versions. % in this study.  

Finally, the comparison results are shown in Table \ref{tab:gen1} and \ref{tab:gen2}. We can see that the English version of our model achieves very competitive results again on all these three test sets. First, when compared with the models that are retrained by the \emph{data augmentation} based method,   we can see that on both SQuAD (Non-Adv-Paraphrased) and  SQuAD (Adv-Paraphrased), the performance of our model is far better than the models like DrQA or BiDAF. And the results of our model are also better than the data augmentation retrained BERT. 
Second, when compared with the models that could not use the additional training data either, we can see that on both SQuAD (Non-Adv-Paraphrased) and  SQuAD (Adv-Paraphrased), our model achieves much competitive results: its results are the best or very close to the best. Third,  \cite{gan2019improving} have  pointed out that although the  {\em data augmentation} based method is usually helpful for models achieve better results  on the paraphrased test sets, it always causes a negligible drop to the performance of models on the original development set. However, from the results in  Table \ref{tab:gen2} we can see that  on SQuAD (dev), a common MRC dataset,   our model achieves better results than all the compared baselines. These results demonstrate an important merit of our model that although it is designed for addressing the robustness issues, there is   almost no any negative affects to  the performance when it is used to handle common MRC datasets. In a word,  our model is a very strong  and  is competent for diverse application scenarios. 

Furthermore, as mentioned above that \cite{hsu-etal-2019-zero} leverage the pre-trained  multi-BERT in cross-lingual zero-shot reading comprehension, which has a similar scenario settings to the robustness issues here because both of them aim to evaluate the performance of models when the distributions of training sets and testing sets are different. Thus there would be a concern  that whether  multi-BERT would also perform well on addressing the  robustness issues in MRC. To answer this concern, we further conduct experiments to demonstrate the performance of  multi-BERT\footnote{\url{https://github.com/google-research/bert}} on these SQuAD-related  datasets. Specifically, we fine-tune  multi-BERT on SQuAD (training), and test it on the mentioned test sets. Here we think it is fair to compare  multi-BERT with our model because both models use  SQuAD (training)  and some multi-lingual resources.  Finally, the results are shown in Table \ref{tab:multibert}. We can see that our model achieves far better results than  multi-BERT on all datasets.  These results indicate that a straightforward multilingual  language model based method could not address the robustness issues in MRC well.

%, datasets are more Their scerio is the same as  our first kind of extreme evaluation. In fact, the first kind of our extreme evaluation can be viewed as a kind of zero-shot evaluation.  But it is different from the second kind of extreme evaluation. 

%It should be noted that although the zero-shot MRC  explored by ~\cite{hsu-etal-2019-zero} is similar to the setting of our first kind of extreme evaluation, we do not take the multi-BERT as a baseline in this extreme evaluation. This is mainly due to the reason that we could not provide a reasonable platform for comparison: if we fine-tune the multi-BERT on DuReader (robust) but test it on the mentioned SQuAD-related dataset, it would be unfair for the multi-BERT model because our model uses the SQuAD (training) in its multi-language learning while multi-BERT not. 

\begin{table*}[t]
	\centering
	%	\makebox[\linewidth]
	\caption{Results  of multi-BERT. P-Questions and A-Questions refer to the paraphrased questions in SQuAD (Non-Adv-Paraphrased) and adversarial questions in SQuAD (Adv-Paraphrased) respectively.  }
	\label{tab:multibert}%
	
	\begin{tabular}{lllllll}
		\toprule
		\multirow{2}[4]{*}{Model} & \multicolumn{2}{c}{P-Questions} & \multicolumn{2}{c}{A-Questions} & \multicolumn{2}{c}{SQuAD (dev)}  \\
		%		\multicolumn{cols}{pos}{text}
		\cmidrule{2-7}          & \multicolumn{1}{l}{EM} & \multicolumn{1}{l}{F1} & \multicolumn{1}{l}{EM} & \multicolumn{1}{l}{F1}  & \multicolumn{1}{l}{EM} & \multicolumn{1}{l}{F1}\\
		\midrule
		\multicolumn{1}{l}{multi-BERT}&76.65&83.94&58.93&66.26&81.39&88.57  \\
		
		\hline
		%			Our Model (Single, \emph{Before}) &76.9 &85.87&56.79&62.22&78.42&87.35  \\
		Our Model  &{\bf 82.96} &{\bf 89.78}&{\bf 76.79}&{\bf 80.92}&{\bf 85.50}&{\bf 92.46}  \\
		\hline
	\end{tabular}

\end{table*}%

\section{Conclusions}
In this study, we propose an \emph{understanding-oriented} MRC model that can  well address the issues of over sensitivity, over stability, and generalization.  We conduct extensive experiments to evaluate it  on three benchmark MRC robustness datasets.  Experimental results show that it  achieves consistent better results   not only  on all of  these robustness MRC datasets, but also on some common MRC datasets. Even on some extreme and unfair evaluations,  it  still achieves much better results. % on both kinds of  datasets. 
%Furthermore, on a common MRC dataset, our model achieves much better results again. 

The main novelties of our model is summarized as follows. 
First,  to the best of our knowledge, this is the first work that  systematically  addresses all three kinds of  robustness issues simultaneously from the model level. 
Second,  we propose a \emph{memory-guided multi-head attention} method that can  mine better  interactions between questions and passages. 
Third, we propose a multi-task and  multi-language learning based  method to integrate the NLI task and the multi-language MRC task together, which is proven to be much effective for addressing  the robustness  issues in MRC. % well.

\begin{acks}
This work is supported by the National Science and Technology Major Project (J2019-IV-0002-0069), the National Natural Science Foundation of China (No.61572120),  and the Fundamental Research Funds for the Central Universities (No.N181602013).
\end{acks}

\bibliographystyle{ACM-Reference-Format}
\bibliography{sample-base}


\end{document}